\documentclass[11pt]{article}

\usepackage{acl}

\usepackage{times}
\usepackage{latexsym}

\usepackage[T1]{fontenc}

\usepackage[utf8]{inputenc}

\usepackage{microtype}

\usepackage{inconsolata}

\usepackage{graphicx}
\usepackage{multirow}
\usepackage{booktabs}
\usepackage{amsmath}
\usepackage{amssymb}
\usepackage{algorithmic}
\usepackage{placeins}
\usepackage{float}
\title{Beyond the Capability Boundary: Zeroth-Order Optimization for Self-Evolving LLM Agents}

\author{
Bingzhen Liu$^{1,2,*}$, Xiaomeng Fan$^{1,2,*}$, Yuwei Wu$^{1,2}$, Zhi Gao$^{1,2}$, Mingyang Gao$^{1,2}$,\\
\textbf{Chuanhao Li}$^{3}$, \textbf{Yunde Jia}$^{1,2}$\\[3pt]
\parbox{0.94\textwidth}{\centering\normalfont
$^{1}$ Beijing Key Laboratory of Intelligent Information Technology,\\
School of Computer Science \& Technology, Beijing Institute of Technology\\
$^{2}$ Guangdong Laboratory of Machine Perception and Intelligent Computing,\\
Shenzhen MSU-BIT University\\
$^{3}$ Alaya Lab
}
}
\begin{document}
\maketitle

\begingroup
\renewcommand{\thefootnote}{}
\footnotetext{
$^{*}$Equal contribution.
}
\endgroup

\begin{abstract}

Self-evolving methods improve the capabilities of LLM agents by sampling trajectories from the underlying LLMs and learning from these trajectories. However, these methods struggle to learn beyond the inherent capability boundary of the agents, since the agents cannot sample correct trajectories on difficult examples for further improvements.
In this paper, we propose a zeroth-order self-evolution framework that enables agents to learn beyond their capability boundary by perturbing LLM parameters to adapt to difficult examples without any trajectory annotations.
Specifically, we perturb LoRA parameters of LLMs, run the agent, compute the losses under the perturbed and original parameters, and use the loss difference to estimate gradients and further update the LoRA parameters.
We sample trajectories using the updated LLMs for supervised fine-tuning to break through the capability boundary of the agents, forming a closed self-evolution loop.
We introduce a parallel perturbation inference mechanism and an adaptive lookup mechanism to reduce time consumption in zeroth-order optimization, with an answer perplexity loss that provides smooth and stable zeroth-order loss values.
Experiments on multiple deep research benchmarks show that our method obtains substantially more successful trajectories and consistently outperforms strong baselines, especially on difficult examples.
The code and released artifacts are available at
\url{https://github.com/hidk1911/ZOForLLMAgents}.

\end{abstract}

\section{Introduction}



Large language model (LLM) agents have shown strong potential in complex information-seeking and reasoning tasks, such as Deep Research \citep{yao2023react,schick2023toolformer,qin2024toolllm}. This task requires agents to retrieve information, integrate evidence, reason over intermediate results, and generate a final answer \citep{nakano2021webgpt,yao2023react}. 
Since annotating high-quality intermediate reasoning trajectories is expensive for training, self-evolving LLM agents have emerged and rapidly attracted increasing attention  \citep{shinn2023reflexion,madaan2023selfrefine,gou2024critic}, which sample reasoning trajectories by themselves and use the trajectories for further training via supervised fine-tuning (SFT) or reinforcement learning (RL)~\citep{wu2026webdancer,sun2025simpledeepsearcher}. However, existing self-evolving LLM agents struggle to learn beyond their inherent capability boundary. On difficult examples, the agents cannot sample correct trajectories for further improvement.

\begin{figure}[t!]
    \centering
    \includegraphics[width=0.95\columnwidth]{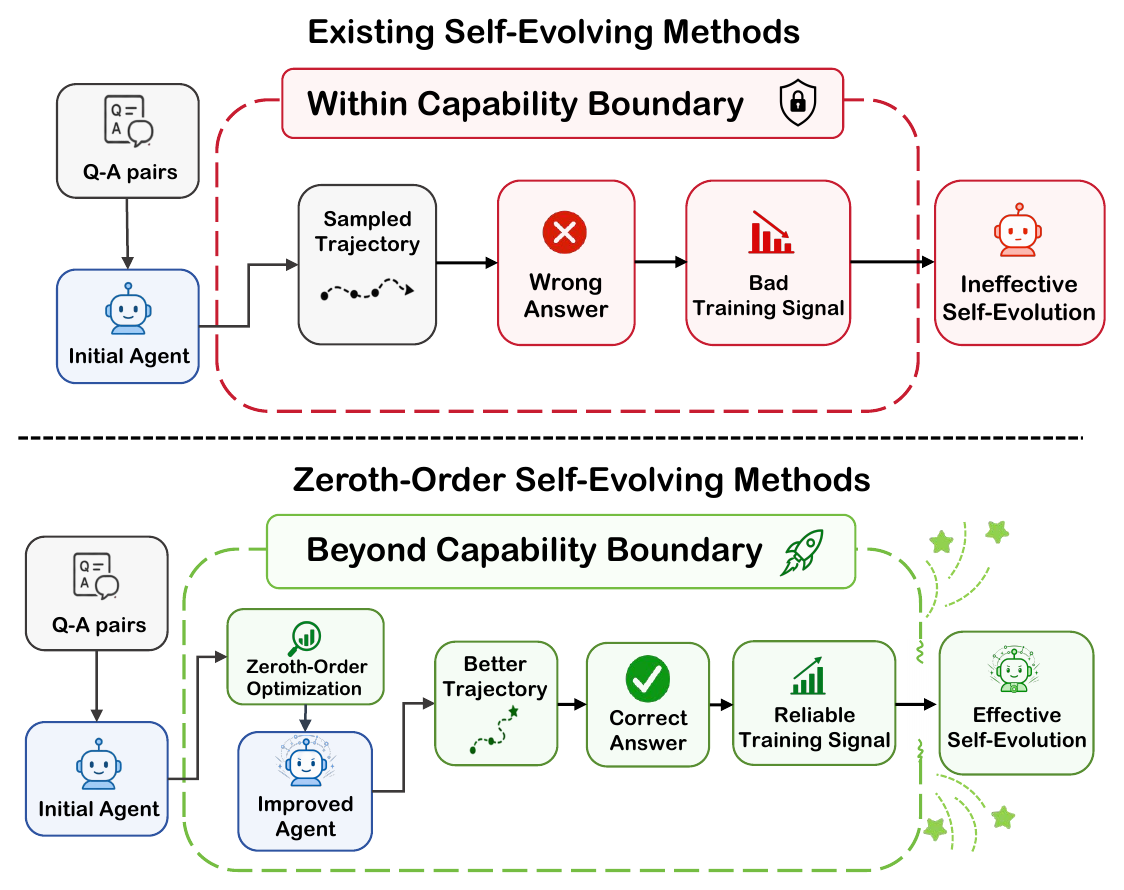}
    \vspace{-0.5em}
    \caption{Comparison between existing self-evolving methods and our zeroth-order self-evolving method.}
    \label{fig:evolution_comparison}
    \vspace{-0.8em}
\end{figure}

In this paper,  we propose a zeroth-order self-evolution framework that enables agents to learn beyond their capability boundary without annotated trajectories. Specifically, for each difficult instance, we attach instance-specific LoRA modules and optimize them using zeroth-order estimation. We randomly perturb the LoRA parameters, compute the final-answer losses under both the original and perturbed parameters, and use the resulting loss differences to construct gradient estimates for updating the LoRA modules.
Since this process only relies on answers and requires no annotated trajectories, it enables optimization without process supervision.
The agent equipped with the optimized LoRA module is capable of generating successful trajectories. 
These newly discovered successful trajectories then serve as high-quality data for subsequent SFT to push the agent beyond its capability boundary, forming a closed self-evolution loop, as shown in Figure~\ref{fig:evolution_comparison}. 

To realize this goal, we need to address two key challenges. First, agentic trajectory sampling is highly expensive. Trajectories involve multiple rounds of tool use and environment interaction, such as retrieval, browsing, filtering, and synthesis. These operations are time-consuming and sensitive to tool instability, latency, and environmental variation. Moreover, since zeroth-order optimization requires evaluating multiple random perturbations, it requires repeatedly sampling full agent trajectories, further introducing prohibitive sampling overhead. Second, without intermediate process (token-level)  supervision, direct final-answer rewards, such as answer similarity or correctness signals, are often noisy and sparse, leading to unstable optimization.

To address the first challenge, we design a parallel perturbation inference mechanism. Since different perturbation branches share the same backbone computation, the expensive backbone output only needs to be computed once, while multiple lightweight LoRA perturbations can be evaluated in parallel. This significantly accelerates zeroth-order optimization with limited extra memory cost. 
We further design an adaptive lookup mechanism that caches recent tool-call results and adaptively uses exact or similarity-based matching according to query type, reducing repeated tool calls and compressing sampling time. 

To address the second challenge, we introduce a answer perplexity loss as a process-level surrogate signal derived only from query and answer pairs. After the model generates a reasoning trajectory, we condition on the trajectory to compute the negative log-likelihood of the reference answer, and use it as the scalar loss for zeroth-order optimization. Compared with sparse final-answer rewards, this loss provides smoother and more discriminative feedback for different trajectories, improving optimization stability.

Experiments on multiple deep research benchmarks show that our method consistently improves final task performance over strong baselines. Compared with existing deep-research agents that rely on trajectories annotated by stronger models such as GPT, our agent achieves better performance, demonstrating that our method can break through the capability boundary of the agent. 
Further analysis shows that our method discovers substantially more successful trajectories, particularly on difficult examples where the original model struggles, further validating its effectiveness.


Our contributions are summarized as follows. 
\begin{itemize}
    \item  We propose a zeroth-order self-evolution method for LLM agents, which enables agents to learn beyond their capability boundary  by perturbing LLM parameters to adapt to difficult examples without any trajectory annotations.
    \item We introduce  a parallel perturbation inference mechanism, an adaptive lookup mechanism, and an answer perplexity loss to improve efficiency and optimization stability.
    \item We conduct systematic experiments to verify the effectiveness of our method and release the sampled high-quality trajectories and trained agents to facilitate future research.
\end{itemize}

\section{Related Work}



\subsection{Self-Evolving Agent}

Self-evolution refers to an iterative process in which a system improves by collecting, refining, and reusing its own experience \citep{tao2024surveyselfevolution}.  Reflexion first developed this idea to LLM agents by using task feedback to guide later agent trials \citep{shinn2023reflexion}. Subsequently, self-evolving methods are developed along two lines, \textit{i.e.}, non-parametric and training-based methods.  Non-parametric methods improve the agent during inference by reusing feedback, memory, or tool-based corrections, such as  Self-Refine \citep{madaan2023selfrefine} using self-generated feedback to revise model outputs, CRITIC \citep{gou2024critic} correcting responses with external tools, and ExpeL \citep{zhao2024expel} storing past task experience as reusable knowledge. Training-based methods further convert  self-generated experience into model updates. WebDancer~\cite{wu2026webdancer}, SimpleDeepSearcher~\citep{sun2025simpledeepsearcher}, and SE-Agent~\citep{guo2026se} construct and improve self-generated trajectories for SFT or RL, and WebEvolver~\citep{fang2025webevolver} introduces a co-evolving world model to provide richer experience.
Existing methods depend on the initial agent to sample trajectories, which is bounded by its capability boundary. Our method uses zeroth-order optimization to reshape the policy, enabling the model to discover trajectories beyond its capability boundary.


\subsection{Zeroth-order Optimization}


Zeroth-order optimization estimates gradients from loss values, making it suitable when gradients are inaccessible or costly~\citep{sun2022blackbox}. Zeroth-order methods mainly follow two directions. Variance-reduction methods aim to stabilize gradient estimations, such as SZVR-G~\cite{liu2018stochastic} and ZO-SVRG \citep{liu2018zeroth}. Acceleration methods aim to improve query efficiency and convergence speed, such as random gradient-free minimization~\cite{nesterov2017random} and ZO-AdaMM \citep{chen2019zoadamm}.

Recently, zeroth-order optimization has been extended to LLMs to reduce training time and memory cost by  avoiding backpropagation~\citep{malladi2023mezo}.
Subsequent studies aim to improve the practicality for LLMs. ZO-LLM~\citep{liu2024revisiting} introduces block-wise zeroth-order updates to improve stability,  and LOZO~\citep{chen2025enhancing} exploits low-rank structures to reduce the cost and variance of zeroth-order estimation. 
However, existing zeroth-order methods have not been studied for LLM agents, which is challenging since  both the trajectory generation and sparse supervision in LLM agents resulting in expensive trajectory sampling and unstable optimization. To fill this gap, we design the adaptive lookup mechanism and answer perplexity loss  for  efficiency and stability.


\section{Method}
\begin{figure*}[t]
    \centering
    \includegraphics[
        width=\textwidth,
        trim=20 110 20 25,
        clip
    ]{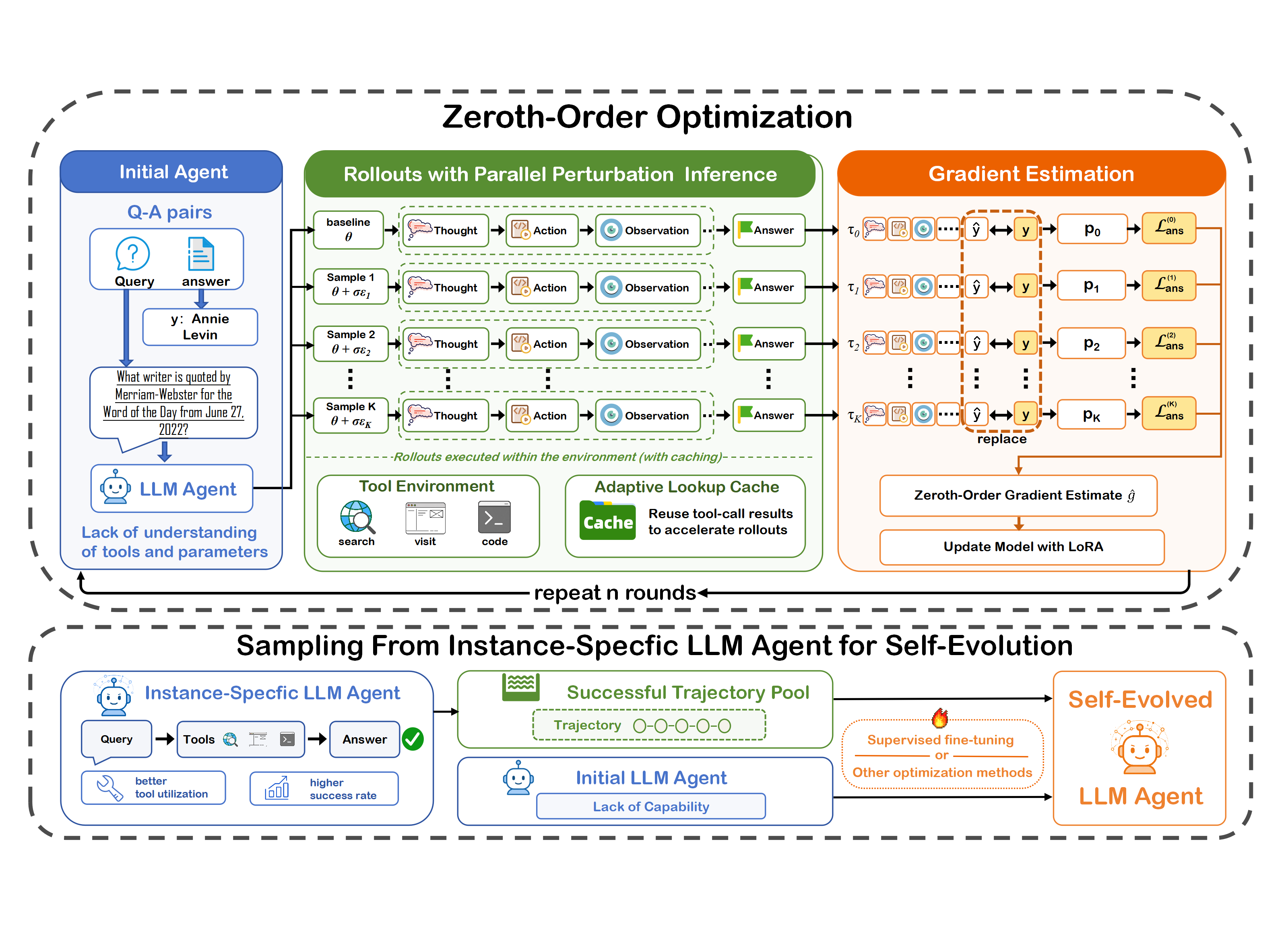}
    \caption{Overview of the proposed zeroth-order self-evolution pipeline. The method actively searches for successful agent trajectories around the current policy by perturbing LoRA parameters and evaluating complete rollouts with answer-only supervision.}
    \label{fig:main_pipeline}
\end{figure*}

We present the background of the proposed method.
Then, we present the formulation of the proposed zeroth-order method for self-evolving LLM agents, which can learn beyond the  capability boundary of agents. 
We introduce a parallel perturbation inference mechanism and an adaptive lookup mechanism to reduce time consumption, and an answer perplexity loss to improve optimization stability.
The overall process is summarized in Figure~\ref{fig:main_pipeline}.

\subsection{Background}

\paragraph{LLM Agents.}

We denote the LLM agent as an policy $\boldsymbol{\pi_{\theta}}$, where $\boldsymbol{\theta}$ denotes parameters. The agent interacts with a tool environment $\mathcal{T}$ consisting of search, webpage visiting, and code execution. At step $t$, the interaction history is
\vspace{-1.5mm}
\begin{equation}
    h_t=(q,r_1,a_1,o_1,\ldots,r_{t-1},a_{t-1},o_{t-1}),
\end{equation}
where $r_t$ is the reasoning text, $a_t$ is an action, and $o_t$ is the returned observation. 
For a query $q$, a complete rollout is denoted as
\vspace{-1.5mm}
\begin{equation}
    \tau=(r_1,a_1,o_1,\ldots,r_T,a_T,o_T,\hat{y}),
\end{equation}
where $\hat{y}$ is the model-generated final answer. The trajectory distribution is computed as
\vspace{-1.5mm}
\begin{equation}
    p_{\boldsymbol{\theta}}(\tau\mid q,\mathcal{T})
    =
    \prod_{t=1}^{T}
    \boldsymbol{\pi_{\theta}}(r_t,a_t\mid h_t)
    p_{\mathcal{T}}(o_t\mid a_t,h_t).
\end{equation}
The environment transition $p_{\mathcal{T}}$ includes multiple tool actions. Since actions are discrete and the environment is non-differentiable, the mapping from $\boldsymbol{\theta}$ to $\tau$ is a black-box stochastic process.

\paragraph{Self-Evolution for LLM Agents.}
A self-evolving LLM agent updates itself solely based on query $q$ and answer pairs $y$ (QA pairs), without intermediate reasoning trajectories. 
The agent should acquire trajectories $\tau$ from its own interaction with $p_{\mathcal{T}}$~\citep{xi2025agentgym,zhang2025evolvesearch,he2025openwebvoyager}. By denoting the training set as $\mathcal{D}=\{(q_i,y_i)\}_{i=1}^{N}$, the  objective is 
\vspace{-1.5mm}
\begin{equation}
    F(\boldsymbol{\theta})
    =
    \mathbb{E}_{(q,y)\sim\mathcal{D}}
    \mathbb{E}_{\tau\sim p_{\boldsymbol{\theta}}(\cdot\mid q,\mathcal{T})}
    \left[
        \ell(\boldsymbol{\theta};q,\tau,y)
    \right],
    \label{eq:blackbox_objective}
\end{equation}
where $\ell$ denotes the loss of SFT or RL. 
In SFT, self-sampled trajectories are used to supervise intermediate processes and compute gradients. In RL, these are used to estimate rewards and update the policy.

\paragraph{Analysis of Self-Evolution methods for LLM Agents.}

Under outcome-only self-evolution, successful trajectories are ultimately required to update the shared agent through trajectory-level self-training. When the current policy samples only failed trajectories, treating these trajectories as positive supervision may reinforce incorrect reasoning, query formulation, and tool-use behaviors. Consequently, methods that rely solely on successful trajectories already reachable by the current policy may remain confined to the agent's current trajectory-discovery boundary.

 Specifically, the gradient of $F(\boldsymbol{\theta})$ is decomposed as
\begin{equation}
\begin{aligned}
\nabla_{\boldsymbol{\theta}}
F(\boldsymbol{\theta})
={}&
\boldsymbol{g}_{\mathrm{fixed}}
+
\boldsymbol{g}_{\mathrm{traj}},
\\
\boldsymbol{g}_{\mathrm{fixed}}
={}&
\mathbb{E}_{\tau}
\left[
\nabla_{\boldsymbol{\theta}}
\ell
(\boldsymbol{\theta};q,\tau,y)
\right],
\\
\boldsymbol{g}_{\mathrm{traj}}
={}&
\mathbb{E}_{\tau}
\left[
\ell
(\boldsymbol{\theta};q,\tau,y)
\right.
\\[-1mm]
&\qquad\left.
\cdot
\nabla_{\boldsymbol{\theta}}
\log
p_{\boldsymbol{\theta}}
(\tau\mid q,\mathcal{T})
\right].
\end{aligned}
\label{eq:trajectory_objective_gradient}
\end{equation}
Here, $\boldsymbol{g}_{\mathrm{fixed}}$ is the explicit gradient of the answer loss when the sampled trajectory is treated as fixed, whereas $\boldsymbol{g}_{\mathrm{traj}}$ captures how parameter changes reshape the trajectory distribution. Therefore, the full gradient contains both an answer-level optimization signal and a trajectory-distribution-level optimization signal.

Given a fixed failed trajectory $\tau^{-}$, a conventional fixed-trajectory first-order update only computes
\begin{equation}
\boldsymbol{g}_{\mathrm{FO}}
=
\nabla_{\boldsymbol{\theta}}
\ell
(\boldsymbol{\theta};q,\tau^{-},y),
\label{eq:fixed_trajectory_fo}
\end{equation}
which corresponds to a single-sample estimate of $\boldsymbol{g}_{\mathrm{fixed}}$.
This gradient increases the likelihood of the reference answer under the existing failed context. However, the update does not account for how changing $\boldsymbol{\theta}$ affects subsequent reasoning decisions, query formulation, tool selection, or environment interaction. Therefore, its gradient provides few trajectory-level information for discovering a successful trajectory.


RL methods also cannot provide effective updates when the current policy cannot sample successful trajectories.
RL methods account for changes in the trajectory using the estimator
\begin{equation}
    \widehat{\boldsymbol{g}}_{\mathrm{RL}}
    =
    \frac{1}{M}
    \sum_{i=1}^{M}
    \left(
        R(\tau_i)-b
    \right)
    \nabla_{\boldsymbol{\theta}}
    \log
    p_{\boldsymbol{\theta}}
    (\tau_i\mid q,\mathcal{T}),
    \label{eq:score_function_rl}
\end{equation}
where $R(\tau_i)$ is the trajectory-level reward and $b$ is a baseline. For difficult examples, if all sampled trajectories fail and receive the same correctness reward $R(\tau_i)=b$, resulting in $\widehat{\boldsymbol{g}}_{\mathrm{RL}}=\boldsymbol{0}$.
In conclusion, existing self-evolving methods restrict the generated trajectories to the model's current capability boundary. 




\paragraph{Zeroth-order Optimization.}

Zeroth-order optimization aims to optimize a loss function when gradients are impractical to compute. 
Instead of relying on backpropagation, it treats the loss as a black box and estimates the gradient only from function evaluations. 
The core idea is to perturb $\boldsymbol{\theta}$ along a random direction $\boldsymbol{u}$, observe the change in the loss values, and use this finite-difference signal to construct an approximate gradient for updating the model, where the gradient is computed as 
\vspace{-1mm}
\begin{equation}
    \hat{g}(\boldsymbol{\theta})
    =
    \frac{F(\boldsymbol{\theta}+\mu \boldsymbol{u})-F(\boldsymbol{\theta})}{\mu} \boldsymbol{u},
\end{equation}
where $F(\cdot)$ denotes a loss function, $\boldsymbol{u}$ is sampled from a standard Gaussian, and $\mu$ is the smoothing radius. 
With such estimated gradients and given a learning rate $\eta$, parameters $\boldsymbol{\theta}$ is updated as, 
\vspace{-1mm}
\begin{equation}
    \boldsymbol{\theta}_{t+1} = \boldsymbol{\theta}_t - \eta \boldsymbol{\hat{g}}(\boldsymbol{\theta}_t),
\end{equation}
thereby enabling optimization under black-box settings where only loss values are required.

\subsection{Formulation}

We use query and answer pairs as feedback and optimize an instance-specific LoRA module for each instance. QA-only supervision can not provide gradients, making direct parameter updates infeasible. 
We thus adopt zeroth-order optimization to perform black-box parameter updates based solely on loss values.



For each instance $(q_i,y_i)$, the instance-specific LoRA module is denoted as $\boldsymbol{w}_i =  [\boldsymbol{A}_i, \boldsymbol{B}_i]$. To optimize $\boldsymbol{w}_i$, we sample  $K$ random perturbation directions. 
For the $k$-th perturbation, the perturbed policy  is denoted as $p_{ \boldsymbol{\theta}_{i,k}^{p}}(\cdot)$, where the perturbed parameters  $ \boldsymbol{\theta}_{i,k}^{p}$ is computed as 
\vspace{-1.5mm}
\begin{equation}
\label{eqaution:parameter_perturb}
    \begin{aligned}
       \boldsymbol{\theta}_{i,k}^{p} = \boldsymbol{\theta}+ (\boldsymbol{B}_i +  \sigma\boldsymbol{\epsilon}^{B}_{i,k}) 
(\boldsymbol{A}_i +  \sigma\boldsymbol{\epsilon}^{A}_{i,k}),
    \end{aligned}
\end{equation} where $\sigma$ controls the perturbation scale. $\boldsymbol{\epsilon}^{B}_{i,k}$ and $\boldsymbol{\epsilon}^{A}_{i,k}$ are perturbations for parameters $\boldsymbol{B}_i$ and $\boldsymbol{A}_i$, respectively. The perturbed policy interacts with the same tool environment and generates a complete trajectory via 
\vspace{-1.5mm}
\begin{equation}
    \tau_{i,k}^{+}
    \sim
    p_{ \boldsymbol{\theta}_{i,k}^{p}}(\cdot \mid q_i,\mathcal{T}).
\end{equation}
The corresponding loss of the perturbed model  is computed by
\vspace{-1.5mm}
\begin{equation}
    \ell_{i,k}^{+}
    =
    \ell_{\mathrm{ans}}
    (  \boldsymbol{\theta}_{i,k}^{p};q_i,\tau_{i,k}^{+},y_i),
\end{equation}
where $\ell_{\mathrm{ans}}$ denotes a loss function computed solely from the final answer.
We also need to utilize the current policy $\boldsymbol{\pi_{\theta}}$  performs a complete rollout and produces a baseline trajectory $\tau_i^0 \sim p_{\boldsymbol{\theta}}(\cdot \mid q_i,\mathcal{T})$, and we compute the baseline loss on the un-perturbed policy as
\vspace{-1.5mm}
\begin{equation}
    \ell_i^0 =
    \ell_{\mathrm{ans}}(\boldsymbol{\theta};q_i,\tau_i^0,y_i).
\end{equation}

The zeroth-order gradient of $w_i$ is estimated by comparing the perturbed loss with the baseline loss, 
\vspace{-1.5mm}
\begin{equation}
   \begin{aligned}
        \hat{g}(\boldsymbol{A}_{i})
    =
    \frac{1}{BK}
    \sum_{i=1}^{B}
    \sum_{k=1}^{K}
    \frac{
        \ell_{i,k}^{+}
        -
        \ell_i^0
    }{\sigma}
    \boldsymbol{\epsilon}_{i,k}^{A},
    \\
     \hat{g}(\boldsymbol{B}_{i})
    =
    \frac{1}{BK}
    \sum_{i=1}^{B}
    \sum_{k=1}^{K}
    \frac{
        \ell_{i,k}^{+}
        -
        \ell_i^0
    }{\sigma}
    \boldsymbol{\epsilon}_{i,k}^{B}.
   \end{aligned}
    \label{eq:zo_pipeline_estimator}
\end{equation}
When a perturbation leads to a lower loss, the update moves the LoRA parameters toward that direction. When a perturbation increases the loss, the update moves away from it. We then update the LoRA parameters $\boldsymbol{w}_i$ with Adam.
By the zeroth-order optimization process, the agent equipped with the optimized LoRA module is encouraged to generate successful reasoning trajectories that are beyond the  capability boundary.  Trajectories with correct final answers or sufficiently low answer loss are added to a high-quality trajectory buffer. This buffer provides the training data for the subsequent SFT, forming a closed self-evolution loop.

\subsection{Parallel Perturbation Inference}

In zeroth-order optimization, the model usually needs to perform forward evaluation under multiple perturbation directions. If we directly perturb the full model parameters, each perturbation requires a separate forward pass through the whole backbone, causing the computation cost to grow linearly with the number of perturbations.

\begin{figure}[t]
    \centering
    \includegraphics[width=\columnwidth]{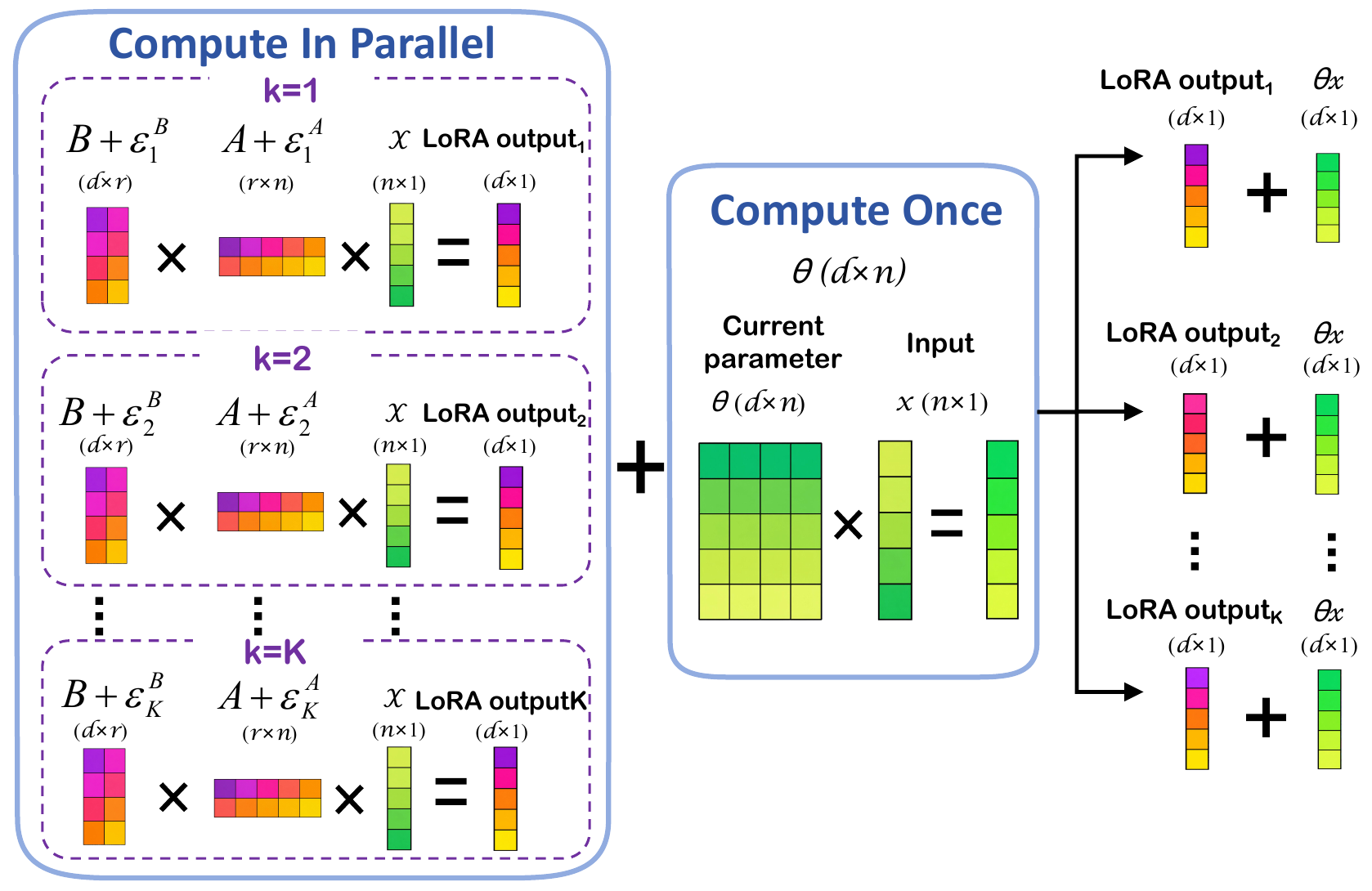}
    \vspace{-0.5em}
    \caption{Illustration of parallel perturbation inference. The backbone output is computed once and shared across perturbation branches, while multiple LoRA perturbation branches are computed in parallel.}
    \label{fig:lora_parallel}
    \vspace{-0.8em}
\end{figure}

For improving efficiency, we  design a parallel perturbation inference that perturbs LoRA parameters in a parallel way. 
With the perturbed parameters in Eq.~\eqref{eqaution:parameter_perturb}, as to the input $\boldsymbol{x}$, for the $k$-th perturbation, the model output can be written as
\vspace{-1.5mm}
\begin{equation}
\label{equation:output_perturb}
    \begin{aligned}
        \boldsymbol{h}_k = \boldsymbol{\theta x} + (\boldsymbol{B}+\boldsymbol{\epsilon}_k^B)(\boldsymbol{A}+\boldsymbol{\epsilon}_k^A)\boldsymbol{x}.
    \end{aligned}
\end{equation}
 We observe that when perturbations only have effects in the LoRA branch, different perturbed models share the same backbone computation $\boldsymbol{\theta x}$. Therefore, we only compute the backbone output $\boldsymbol{\theta x}$ once and share it across all perturbation branches, which is illustrated in Figure~\ref{fig:lora_parallel}. In detail, for $1-k$ perturbations, we in parallel compute 
 the right term in Eq.~\eqref{equation:output_perturb}, \textit{i.e.},
 $(\boldsymbol{B}+\boldsymbol{\epsilon}_1^B)(\boldsymbol{A}+\boldsymbol{\epsilon}_1^A)\boldsymbol{x}, ..., (\boldsymbol{B}+\boldsymbol{\epsilon}_K^B)(\boldsymbol{A}+\boldsymbol{\epsilon}_K^A)\boldsymbol{x}$ are computed in parallel. 
 Then, we add $\boldsymbol{\theta x}$ to these terms as $\boldsymbol{\theta x}+(\boldsymbol{B}+\boldsymbol{\epsilon}_1^B)(\boldsymbol{A}+\boldsymbol{\epsilon}_1^A)\boldsymbol{x}, ..., \boldsymbol{\theta x}+(\boldsymbol{B}+\boldsymbol{\epsilon}_K^B)(\boldsymbol{A}+\boldsymbol{\epsilon}_K^A)\boldsymbol{x}$.
Due to the  low-rank structures of $\boldsymbol{A}, \boldsymbol{B}$, their computation and memory cost are much smaller. In this way, we can evaluate multiple perturbations on top of a single backbone forward computation in parallel, which greatly reduces the sampling time cost of zeroth-order optimization.

\subsection{Adaptive Lookup Mechanism}

Due to environmental and network conditions, tool calls can be quite time-consuming.
Evaluating multiple perturbations for the same question often leads to repeated tool calls, such as similar \texttt{search} queries or visit the same webpages, which makes direct execution of every tool action unnecessarily expensive.


We therefore introduce an adaptive lookup mechanism to reuse tool observations. We maintain a shared pool of recent tool calls and returned results. For \texttt{search}, the lookup uses semantic query matching to handle surface-level variations. For \texttt{visit}, it uses stricter matching based on the URL and visit goal. Matched requests are served from the lookup pool, while unmatched requests are executed externally and registered for future reuse.

This design reduces redundant external interactions without changing the agent rollout format. It substantially lowers the time cost of multi-perturbation rollouts and improves the practicality of our zeroth-order self-evolution pipeline.

\subsection{Answer Perplexity Loss}


Zeroth-order gradient estimation relies on the loss values evaluated under perturbed parameters, making the design of the loss function critical for stability. 
Binary loss and BERT-based similarity loss are discontinuous, leading to noisy gradient estimation that causes the optimization process to oscillate and fail to converge.

We introduce an answer perplexity loss that uses trajectory-conditioned answer likelihood as dense supervision~\citep{acikgoz2026tt}. Given a rollout $\tau$ for query $q$, the final answer is required to appear in the format  $\texttt{<answer>}~\hat{y}~\texttt{</answer>}$.
We locate this answer span and replace the generated answer with the ground-truth answer. We denote the trajectory context generated by our agent as $\tau_{\setminus \mathrm{ans}}$. The ground-truth answer is tokenized as $y=(y_1,\ldots,y_M)$. The introduced answer perplexity loss is
\vspace{-1.5mm}
\begin{equation}
\small
    \ell_{\mathrm{ans}}(\boldsymbol{\theta};q,\tau,y)
    =
    -\frac{1}{M}
    \sum_{j=1}^{M}
    \log
    \boldsymbol{\pi_{\theta}}
    \left(
        y_j
        \mid
        q,\tau_{\setminus \mathrm{ans}},y_{<j}
    \right).
    \label{eq:answer_nll}
\end{equation}

The loss in Eq.~\eqref{eq:answer_nll} is computed only on ground-truth answer tokens. We do not supervise the generated reasoning, actions and observations. 
Instead, the loss evaluates whether the sampled trajectory provides conditions under which the ground-truth answer is easy to predict. 

Strictly, Eq.~\eqref{eq:answer_nll} is a token-normalized negative log-likelihood or cross-entropy loss. The corresponding answer perplexity is
\vspace{-1.5mm}
\begin{equation}
    \mathrm{PPL}_{\mathrm{ans}}(\boldsymbol{\theta};q,\tau,y)
    =
    \exp\left(
        \ell_{\mathrm{ans}}(\boldsymbol{\theta};q,\tau,y)
    \right).
\end{equation}
We optimize the log-space form because it is numerically stable and directly usable for loss-difference estimation. A lower loss means that the current trajectory makes the reference answer less surprising to the model. Therefore, even an incorrect rollout can still receive useful credit if it retrieves evidence that increases the likelihood of the ground-truth answer.

\section{Experiment}

\subsection{Experimental Setup}

\paragraph{Evaluation Benchmarks.}We evaluate our method on two deep research  benchmarks: GAIA~\citep{mialon2024gaia} and WebWalkerQA~\citep{wu2025webwalker}.
GAIA tests general assistant abilities including reasoning, browsing, multimodal handling, and tool use. WebWalkerQA focuses on website traversal and multi-page information extraction.
These benchmarks are well suited to our study because their multi-step search difficulty provides a direct test of whether our discovered trajectories can produce stronger agents.


\paragraph{Training data.}
Starting from 500 QA pairs from the open-source WebShaper data~\citep{tao2026webshaper}, we filter 304 high-quality QA pairs for training.

\paragraph{Metrics.}We report answer accuracy under the official evaluation protocol. 


\paragraph{Comparison.}
We compare our method with the baseline ReAct framework~\citep{yao2023react}. 
For controlled comparisons, all internal methods use the same backbone family and the same system configuration, including the search API, webpage parser, decoding setting, and tool budget.
We further compare our method with existing deep-research agents, including Search-o1~\citep{li2025searcho1}, WebThinker~\citep{li2026webthinker}, WebDancer~\citep{wu2026webdancer}, SimpleDeepSearcher~\citep{sun2025simpledeepsearcher}, WebSailor~\citep{li2025websailor}, GRPO~\citep{shao2024deepseekmath}, and ARPO~\citep{dong2025arpo}. 




\
\subsection{Main Results
}

Table~\ref{tab:main_results} reports the main controlled results and contextual paper-reported comparisons. 
Results show that our method surpasses the baseline method, demonstrating the effectiveness of our method.

Because the compared deep-research agents rely on more large-scale models (\textit{e.g.} GPT) to generate trajectories,
these agents are easier to obtain strong results than self-evolution agents. Despite this disadvantage, our method still outperforms these agents, demonstrating its superiority and showing that it can break through the capability boundary of self-evolving agents.



\begin{table*}[t]
\centering
\scriptsize
\setlength{\tabcolsep}{2.5pt}
\renewcommand{\arraystretch}{0.94}
\caption{Main results under controlled evaluation and contextual paper-reported comparisons with recent deep-research agents. Paper-reported rows are grouped by framework/search scaffolds and training-based methods.}
\label{tab:main_results}
\resizebox{\textwidth}{!}{
\begin{tabular}{llcccccccccc}
\toprule
\multirow{2}{*}{Backbone} & \multirow{2}{*}{Method}
& \multicolumn{4}{c}{GAIA}
& \multicolumn{4}{c}{WebWalkerQA} \\
\cmidrule(lr){3-6} \cmidrule(lr){7-10}
& & Level 1 & Level 2 & Level 3 & Avg.
& Easy & Medium & Hard & Avg.
& & \\
\midrule
\multicolumn{12}{l}{\textit{Paper-reported framework/search scaffolds}} \\
Qwen-2.5-7B & ReAct~\citep{yao2023react}
& 28.2 & 15.3 & 0.0 & 18.4
& 28.1 & 31.2 & 16.0 & 24.2
\\
Qwen-2.5-7B & Search-o1~\citep{li2025searcho1}
& 23.1 & 17.3 & 0.0 & 17.5
& -- & -- & -- & --
\\
Qwen-3-8B & Search-o1
& 35.9 & 15.4 & 0.0 & 21.4
& 6.7 & 15.5 & 9.7 & 11.5
\\
Qwen-3-8B & WebThinker~\citep{li2026webthinker}
& 43.6 & 11.5 & 0.0 & 22.3
& 6.7 & 13.1 & 16.9 & 13.0
\\
\midrule
\multicolumn{12}{l}{\textit{Paper-reported training-based methods}} \\
Qwen-2.5-7B & WebDancer~\citep{wu2026webdancer}
& 41.0 & 30.7 & 0.0 & 31.0
& 40.6 & 44.1 & 28.2 & 36.0
\\
Qwen-2.5-7B & SimpleDeepSearcher$^{\dagger}$~\citep{sun2025simpledeepsearcher}
& -- & -- & -- & 36.9
& -- & -- & -- & --
\\
WebSailor-7B & WebSailor$^{\dagger}$~\citep{li2025websailor}
& -- & -- & -- & 33.0
& -- & -- & -- & --
\\
Qwen-3-8B & GRPO~\citep{shao2024deepseekmath}
& 48.7 & 25.0 & 8.3 & 32.0
& 24.4 & 33.3 & 26.8 & 29.0
\\
Qwen-3-8B & ARPO~\citep{dong2025arpo}
& 53.9 & 32.7 & 16.7 & 38.8
& 26.7 & 33.3 & 29.6 & 30.5
\\
\midrule
\multicolumn{12}{l}{\textit{Controlled setting}} \\
Qwen-3-4B & ReAct
& 33.3 & 15.1 & 5.0 & 19.6
& 25.0 & 17.5 & 18.3 & 19.5
\\
& Ours-SFT
& \textbf{42.8} & \textbf{24.2} & \textbf{10.0} & \textbf{28.3}
& \textbf{30.0} & \textbf{26.4} & \textbf{29.2} & \textbf{28.2}
\\
\addlinespace[1pt]
Qwen-3-8B & ReAct
& 35.9 & 17.3 & 8.3 & 23.3
& 8.9 & 16.7 & 18.3 & 15.5
\\
& Ours-SFT
& \textbf{56.4} & \textbf{42.3} & \textbf{41.6} & \textbf{47.5}
& \textbf{31.4} & \textbf{36.8} & \textbf{35.2} & \textbf{34.8}
\\
\midrule
\end{tabular}
}
\end{table*}

\subsection{Ablation}

We conduct ablation studies from two perspectives. First, we compare different loss functions  for zeroth-order trajectory discovery. Second, we analyze whether our method can help the agent discover successful trajectories on difficult QA pairs.

\subsubsection{Loss convergence}

We compare three loss functions: BERT-based semantic loss, LLM-as-a-Judge loss, and the proposed answer perplexity loss. The comparison is conducted on an easy example and a hard example 
 with the same question under the same zeroth-order setting. 
Since these three losses have different scales, we normalize their values to the range of 0 to 1 for a fair comparison of their convergence.

\begin{figure}[t]
    \centering
    \begin{minipage}{0.48\columnwidth}
        \centering
        \includegraphics[width=\linewidth]{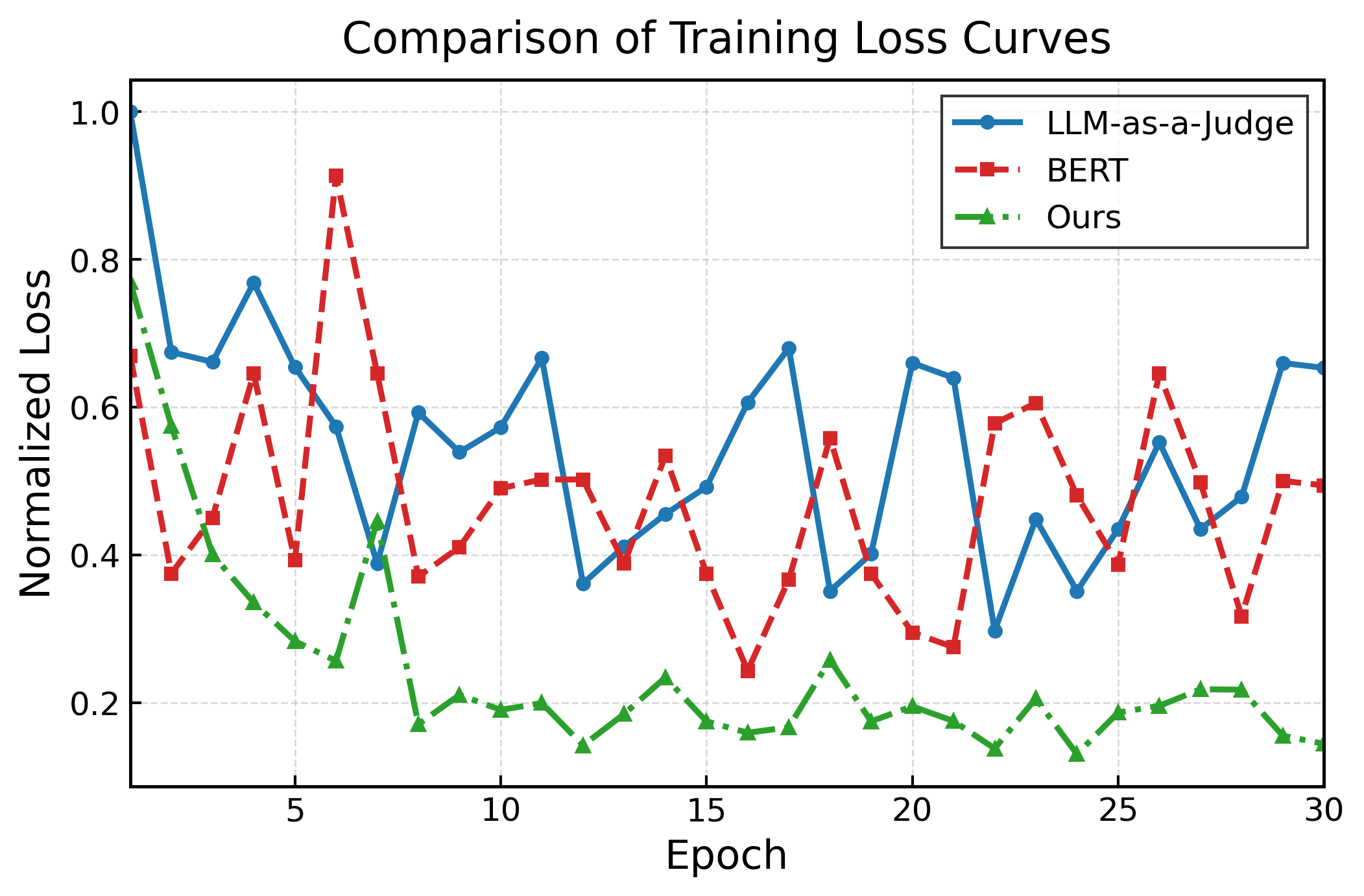}
        \centerline{\scriptsize (a) Results on the easy sample.}
    \end{minipage}
    \hfill
    \begin{minipage}{0.48\columnwidth}
        \centering
        \includegraphics[width=\linewidth]{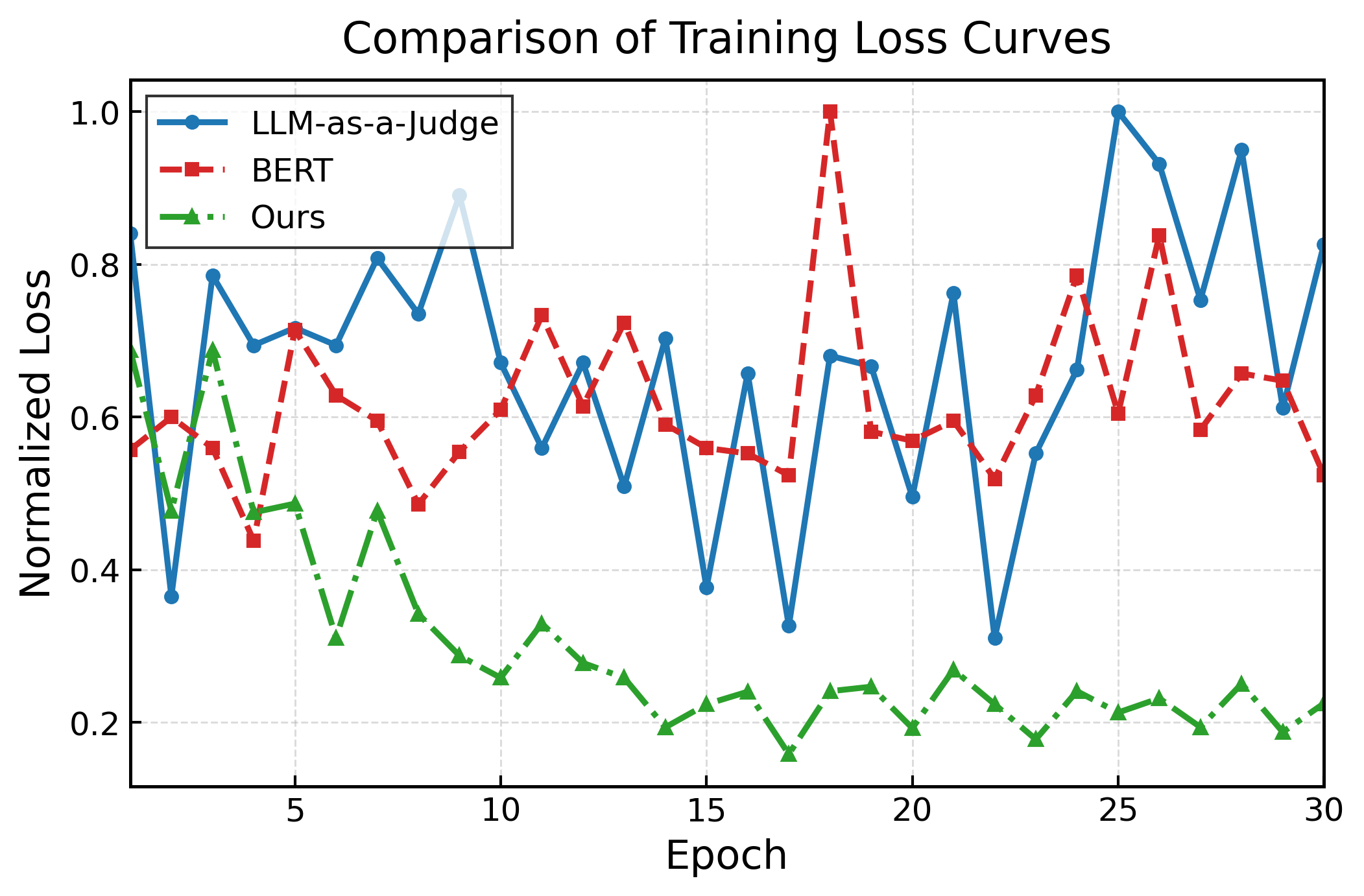}
        \centerline{\scriptsize (b) Results on the hard sample.}
    \end{minipage}
    \caption{Convergence comparison of different loss functions on easy and hard examples.}
    \label{fig:loss_convergence}
\end{figure}

Results are shown in Figure~\ref{fig:loss_convergence}.
On the easy example, all three losses can decrease during zeroth-order optimization, but our loss achieves a lower final loss value and leads to a more stable optimization process. 
On difficult samples, both the LLM-as-a-Judge loss and the BERT-based loss fail to decrease, whereas our loss still converges normally with stable optimization dynamics. 
These results demonstrate that the proposed loss is well suited for zeroth-order optimization and is effective in providing reliable optimization signals.


\subsubsection{Effect on difficult QA pairs}
Among the 304 training examples, the baseline model correctly answers only 67 examples under Pass@1 and fails on the remaining 237, resulting in an initial success rate of 22.0\%. 
After training with the proposed zeroth-order optimization method, the optimized agent correctly answers 164 examples under Pass@1, increasing the success rate to 53.9\% and yielding an improvement of 31.9\%. 
Results demonstrate that agents trained with the proposed method can effectively break through the capability boundary of the initial agent.


We further train the initial agent using the 67 successful trajectories sampled from the initial agent and evaluate the resulting self-evolved agent on the GAIA benchmark. 
As shown in Table~\ref{tab:base_success_vs_ours_gaia_4b}, training on trajectories sampled from the initial model improves over the baseline agent, demonstrating the effectiveness of self-evolution. 
More importantly, our method significantly outperforms the two compared agents, demonstrating that our method can help sample high-quality trajectories and validating its effectiveness.





\begin{table}[t]
\centering
\footnotesize
\setlength{\tabcolsep}{3.5pt}
\renewcommand{\arraystretch}{0.92}
\caption{GAIA accuracy comparison of the agents (Qwen-3-4B as the backbone) trained with different trajectory sources.}
\label{tab:base_success_vs_ours_gaia_4b}
\resizebox{\columnwidth}{!}{
\begin{tabular}{lccccc}
\toprule
Agent. & Level 1 & Level 2 & Level 3 & Avg.  \\
\midrule
Agent w/o training & 33.3 & 15.1 & 5.0  & 19.6  \\
Agent trained on initial trajectories  & 38.1 & 13.6 & 5.0 & 20.4  \\
Agent trained on our trajectories  & \textbf{42.8} & \textbf{24.2} & \textbf{10.0} & \textbf{28.3}  \\
\bottomrule
\end{tabular}
}
\end{table}

\subsubsection{Hyperparameter sensitivity}
We randomly select 50 difficult examples and vary one hyperparameter at a time while keeping the remaining settings fixed. We evaluate the number of perturbation directions $K$, perturbation scale $\sigma$, LoRA rank $r$, and learning rate $\eta$. Table~\ref{tab:zo_sensitivity} shows that the number of successful examples changes only slightly over most tested settings. The default configuration is not the only effective choice, suggesting that the trajectory-discovery results are not narrowly tuned to a single hyperparameter setting.

\begin{table}[t]
\centering
\footnotesize
\setlength{\tabcolsep}{4pt}
\renewcommand{\arraystretch}{0.92}
\caption{Hyperparameter sensitivity on 50 difficult examples. We vary one factor at a time; default settings are bolded.}
\label{tab:zo_sensitivity}
\begin{tabular*}{\columnwidth}{@{\extracolsep{\fill}}lcc@{}}
\toprule
Factor & Tested value & Successful instances \\
\midrule
\multirow{4}{*}{$K$}
& 1 & 20 / 50 \\
& \textbf{2} & \textbf{23 / 50} \\
& 4 & 24 / 50 \\
& 6 & 24 / 50 \\
\midrule
\multirow{3}{*}{$\sigma$}
& $5\times10^{-3}$ & 22 / 50 \\
& $\mathbf{1\times10^{-3}}$ & \textbf{23 / 50} \\
& $5\times10^{-4}$ & 21 / 50 \\
\midrule
\multirow{3}{*}{$r$}
& 8 & 20 / 50 \\
& \textbf{16} & \textbf{23 / 50} \\
& 64 & 23 / 50 \\
\midrule
\multirow{3}{*}{$\eta$}
& $5\times10^{-6}$ & 22 / 50 \\
& $\mathbf{1\times10^{-6}}$ & \textbf{23 / 50} \\
& $5\times10^{-7}$ & 19 / 50 \\
\bottomrule
\end{tabular*}
\end{table}

\subsubsection{One-sided versus two-sided estimation}
The two-sided estimator of zero-order gradients evaluates positive and negative perturbations for each direction, whereas the one-sided estimator reuses a shared unperturbed rollout. We compare the two estimators under the same four-rollout budget per round. As shown in Table~\ref{tab:zo_estimator_comparison}, the two-sided estimator solves 24 examples and the one-sided estimator solves 23. Since their trajectory-discovery performance is comparable, we use the one-sided estimator in the main experiments because it supports more perturbation directions with a shared baseline rollout.

\begin{table}[H]
\centering
\footnotesize
\setlength{\tabcolsep}{3pt}
\renewcommand{\arraystretch}{0.92}
\caption{Comparison of one-sided and two-sided estimators on 50 difficult examples under the same rollout budget.}
\label{tab:zo_estimator_comparison}
\resizebox{\columnwidth}{!}{
\begin{tabular}{lcccc}
\toprule
Estimator
& Directions
& Rollouts / direction
& Rollouts / round
& Successful \\
\midrule
One-sided (Ours)
& 3
& 1 + shared baseline
& 4
& 23 / 50 \\
Two-sided SPSA
& 2
& 2
& 4
& \textbf{24 / 50} \\
\bottomrule
\end{tabular}
}
\end{table}

\subsection{Trajectory Quality Analysis}

We further present representative trajectories to show that zeroth-order optimization breaks through the capability boundary  of the agent. We present cases where the baseline fails but the optimized agent succeeds, which is shown in Table~\ref{tab:badcase_qualitative}. Results show that the baseline agent often fails because of its wrong reasoning and tool call. In contrast, the optimized agent makes more purposeful tool calls, reformulates insufficient queries, and grounds its final answer more explicitly in retrieved evidence.  These cases suggest that our method improves trajectory-level competence rather than only changing the final answer distribution.



\begin{table*}[t]
\centering
\scriptsize
\setlength{\tabcolsep}{4pt}
\renewcommand{\arraystretch}{0.95}
\caption{Qualitative comparison between the baseline agent and our evolved agent.}
\label{tab:badcase_qualitative}
\resizebox{\textwidth}{!}{
\begin{tabular}{p{0.12\textwidth}p{0.38\textwidth}p{0.38\textwidth}}
\toprule
Case & Baseline Agent & Ours \\
\midrule
Case 1: Evidence misinterpretation 
\par
\textbf{Question:} Shared model type in Kashyap's and Fader's customer-retention studies.
\par
\textbf{Ground Truth:} \texttt{beta geometric}
&
\textbf{Thinking:} Relies on prior belief and selects ``Markov model'' too early.
\par
\textbf{Action:} \texttt{search} for Kashyap and Fader retention studies.
\par
\textbf{Observation:} Retrieves beta-geometric clues but does not verify them.
\par
\textbf{Error turn:} \textbf{Prior belief overrides retrieved evidence.}
\par
\textbf{Answer:} \texttt{Markov model}.
&
\textbf{Thinking:} Treats ``beta geometric'' as the key candidate from retrieved evidence.
\par
\textbf{Action:} \texttt{search} for Kashyap-Fader studies, then \texttt{search} for beta-geometric retention models.
\par
\textbf{Observation:} Confirms beta geometric evidence from both Kashyap and Fader-related results.
\par
\textbf{Correction:} Verifies the candidate before answering.
\par
\textbf{Answer:} \texttt{beta geometric}. \\
\midrule
Case 2: Weak query formulation 
\par
\textbf{Question:} Horror movie cited by Valentina Re for dream-reality metalepsis.
\par
\textbf{Ground Truth:} \texttt{A Nightmare on Elm Street}
&
\textbf{Thinking:} Considers several dream-reality horror films but lacks a stable evidence target.
\par
\textbf{Action:} \texttt{search} with separated broad terms such as Valentina Re, metalepsis, dream reality, and horror movie.
\par
\textbf{Observation:} Retrieves generic book and horror-film results without identifying the cited movie.
\par
\textbf{Error turn:} \textbf{Guesses from common horror examples instead of refining the search.}
\par
\textbf{Answer:} \texttt{The Others}.
&
\textbf{Thinking:} Focuses on the specific chapter context and the dream-reality metalepsis clue.
\par
\textbf{Action:} \texttt{search} for Valentina Re, World Building, metalepsis, and horror movie; then refines with Valentina Re and metalepsis horror.
\par
\textbf{Observation:} Finds Re's chapter on metalepsis and narrows the dream-reality horror candidate.
\par
\textbf{Correction:} Uses targeted search to resolve the cited film.
\par
\textbf{Answer:} \texttt{A Nightmare on Elm Street}. \\
\bottomrule
\end{tabular}
}
\end{table*}

\subsection{Capability Boundary Analysis}

\subsubsection{Comparison with first-order alternatives}
We randomly select 50 difficult examples and compare the proposed method with two gradient-based alternatives under the same backbone, tools, and rollout budget. The fixed-trajectory first-order (FO) baseline keeps the failed trajectory sampled by the initial policy fixed and backpropagates only the answer loss. The RL baseline regenerates trajectories from the current policy and uses an outcome-level answer reward. As shown in Table~\ref{tab:fo_rl_comparison}, the fixed-trajectory FO baseline improves only one additional example over direct sampling, while RL solves 16 examples. Our method solves 23 examples, showing that loss differences over regenerated complete rollouts provide more useful guidance for trajectory discovery on difficult examples.

\begin{table*}[t]
\centering
\scriptsize
\setlength{\tabcolsep}{4pt}
\renewcommand{\arraystretch}{0.95}
\caption{Comparison with first-order alternatives on 50 difficult examples. All methods use the same backbone, tools, and rollout budget.}
\label{tab:fo_rl_comparison}
\resizebox{\textwidth}{!}{
\begin{tabular}{llll}
\toprule
Method & Trajectory generation & Update signal & Successful instances \\
\midrule
Initial policy
& Direct sampling from the initial policy
& No parameter update
& 6 / 50 \\
Fixed-trajectory FO
& Keep the sampled failed trajectory fixed
& Answer-span gradient on the fixed context
& 7 / 50 \\
RL
& Regenerate trajectories from the current policy
& Outcome-level answer reward
& 16 / 50 \\
Ours
& Regenerate rollouts under perturbed and updated policies
& Loss differences across complete rollouts
& \textbf{23 / 50} \\
\bottomrule
\end{tabular}
}
\end{table*}

Under the same Qwen-3-4B configuration, fixed-trajectory FO requires 21--24~GB of GPU memory and 2.5--3.5 hours of wall-clock time, whereas our ZO implementation requires 12--15~GB and 1.7--2.4 hours. These measurements show the practical benefit of avoiding backpropagation in this configuration, rather than implying that ZO is uniformly faster than first-order optimization.

\subsubsection{Best-of-$N$ trajectory search}
We further compare trajectory-discovery methods under matched complete-rollout budgets. The initial-policy baseline directly samples $N$ trajectories. Random LoRA perturbation uses the same perturbation scale and perturbed rollouts as our method but does not estimate gradients or update parameters. Answer-conditioned trajectory search provides the reference answer as additional context and searches for a supporting trajectory without parameter updates. An example is counted as solved if at least one of the trajectories obtained within the budget is correct. As shown in Table~\ref{tab:bestofn_trajectory_discovery}, our method achieves the highest success rate for every evaluated budget. The advantage becomes larger as the budget increases, indicating that iterative loss-guided updates are more effective than additional sampling or unguided perturbation.

\begin{table*}[t]
\centering
\footnotesize
\setlength{\tabcolsep}{5pt}
\renewcommand{\arraystretch}{0.94}
\caption{Best-of-$N$ trajectory-discovery success rate (\%) under matched rollout budgets. All methods start from the same initial policy.}
\label{tab:bestofn_trajectory_discovery}

\makebox[\textwidth][c]{%
\begin{tabular*}{0.94\textwidth}{@{\extracolsep{\fill}}lccccc@{}}
\toprule
Method & $N=1$ & $N=2$ & $N=4$ & $N=6$ & $N=8$ \\
\midrule
Initial policy
& 22.04 & 26.97 & 29.93 & 31.57 & 32.23 \\

Random LoRA perturbation
& 21.38 & 27.63 & 31.57 & 32.56 & 33.55 \\

Answer-conditioned trajectory search
& 24.67 & 27.63 & 32.56 & 36.51 & 37.82 \\

Zeroth-order optimization (Ours)
& \textbf{26.97}
& \textbf{30.26}
& \textbf{41.19}
& \textbf{50.32}
& \textbf{53.61} \\
\bottomrule
\end{tabular*}%
}

\end{table*}

\subsubsection{Pass@$N$ scaling}
We evaluate Pass@$N$ for $N\in\{1,2,4,6,8\}$, where an example is considered solved if at least one of the $N$ sampled trajectories is correct. On the trajectory-discovery training set, we compare the initial policy with our optimized policy. On GAIA, we additionally include the RL-trained policy. As shown in Table~\ref{tab:passn_scaling}, our method obtains the best result at every sampling budget. The gap remains when $N$ increases, showing that the improvement is not restricted to Pass@1.

\begin{table*}[t]
\centering
\footnotesize
\setlength{\tabcolsep}{5pt}
\renewcommand{\arraystretch}{0.94}
\caption{Pass@$N$ (\%) on the trajectory-discovery training set and GAIA.}
\label{tab:passn_scaling}

\makebox[\textwidth][c]{%
\begin{tabular*}{0.94\textwidth}{@{\extracolsep{\fill}}llccccc@{}}
\toprule
Split & Policy & Pass@1 & Pass@2 & Pass@4 & Pass@6 & Pass@8 \\
\midrule

\multirow{2}{*}{Training}
& Initial policy
& 22.0 & 26.9 & 30.2 & 32.5 & 33.5 \\

& Ours
& \textbf{53.9}
& \textbf{58.5}
& \textbf{61.5}
& \textbf{63.1}
& \textbf{63.8} \\

\midrule

\multirow{3}{*}{GAIA}
& Initial policy
& 19.6 & 25.2 & 28.1 & 30.1 & 31.1 \\

& RL
& 25.2 & 32.0 & 36.9 & 36.9 & 39.8 \\

& Ours
& \textbf{28.3}
& \textbf{34.9}
& \textbf{37.8}
& \textbf{39.8}
& \textbf{41.7} \\

\bottomrule
\end{tabular*}%
}

\end{table*}

We further match the full rollout budget of zeroth-order optimization with $N_{\mathrm{total}}=45$ samples from the unchanged initial policy. On the training set, the initial policy solves 102 examples, while our method solves 194. Among the 97 examples newly solved by our method over the initial Pass@1 result, only 35 examples (36.1\%) can be recovered by Best-of-$N_{\mathrm{total}}$ sampling from the initial policy. The remaining 62 examples are still unsolved after the same sampling budget. Moreover, the initial-policy result already plateaus at Pass@8, with Pass@8 identical to Pass@$N_{\mathrm{total}}$. These results show that additional sampling alone does not account for the newly discovered trajectories under the evaluated budget.

\subsubsection{Optimization trajectory analysis}
For 50 difficult examples, we record the LoRA parameters and answer loss at every optimization round. For each example, we construct a fixed two-dimensional PCA basis from the parameter displacements of both our method and a matched-budget random-perturbation baseline, and project all checkpoints onto this shared basis. Table~\ref{tab:optimization_trajectory} reports a representative trajectory and aggregate diagnostics. On the representative example, our trajectory moves along a consistent direction while the answer loss decreases from 4.333 to 0.024. In contrast, the random baseline ends with a higher loss than its initial value. Across all 50 examples, our updates have higher consecutive-direction cosine similarity and trajectory efficiency, together with a larger loss decrease and more successful examples. These results are consistent with loss-guided optimization rather than unguided random exploration.

\begin{table*}[t]
\centering
\footnotesize
\setlength{\tabcolsep}{4pt}
\renewcommand{\arraystretch}{0.94}
\caption{Low-dimensional analysis of the parameter optimization trajectory. The PCA basis is shared by our method and the matched-budget random-perturbation baseline for each example.}
\label{tab:optimization_trajectory}

\textbf{(a) Shared-PCA projection on a representative difficult example.}
\vspace{2pt}

\begin{tabular*}{0.90\textwidth}{@{\extracolsep{\fill}}rccc|ccc@{}}
\toprule
\multirow{2}{*}{Round}
& \multicolumn{3}{c|}{Ours}
& \multicolumn{3}{c}{Random perturbation} \\
\cmidrule(lr){2-4}
\cmidrule(lr){5-7}
& PC1 & PC2 & Answer loss
& PC1 & PC2 & Answer loss \\
\midrule
0
& 0.000 & 0.000 & 4.333
& 0.000 & 0.000 & 4.333 \\
4
& -0.354 & 0.059 & 2.541
& -0.173 & 0.314 & 3.486 \\
8
& -1.630 & 0.526 & 0.391
& 0.375 & -0.929 & 6.925 \\
12
& -2.819 & 0.928 & 0.454
& 1.339 & -1.733 & 7.376 \\
15
& -7.008 & 1.617 & 0.024
& -0.564 & -3.811 & 5.197 \\
\bottomrule
\end{tabular*}

\vspace{5pt}
\textbf{(b) Aggregate diagnostics over 50 difficult examples.}
\vspace{2pt}

\begin{tabular*}{0.80\textwidth}{@{\extracolsep{\fill}}lcc@{}}
\toprule
Diagnostic & Ours & Random perturbation \\
\midrule
Mean answer-loss decrease from round 0 to 15
& \textbf{74.8\%} & 13.4\% \\

Checkpoints with loss below the initial value
& \textbf{62.3\%} & 46.3\% \\

Mean cosine similarity between consecutive updates
& \textbf{0.85} & 0.02 \\

Mean trajectory efficiency
$\frac{\lVert w_R-w_0\rVert_2}
{\sum_t\lVert w_{t+1}-w_t\rVert_2}$
& \textbf{0.64} & 0.27 \\

Successful examples after 15 rounds
& \textbf{23} & 5 \\
\bottomrule
\end{tabular*}

\end{table*}

\subsection{Generalization and Robustness}

\subsubsection{Non-retrieval and non-verbatim subsets}
We identify 56 GAIA questions that do not require information retrieval and instead mainly involve multi-step reasoning, computation, or direct problem solving. The initial policy solves 5 examples, the RL baseline solves 11, and our method solves 16, corresponding to accuracies of 8.93\%, 19.64\%, and 28.57\%, respectively. We further evaluate 68 examples for which the gold answer does not appear verbatim in any retrieved observation. The initial policy solves 7 examples, while our method solves 21. This result shows that the answer loss remains useful when the agent cannot obtain the answer through direct copying from retrieved text.

\subsubsection{Performence on harder benchmarks}
We evaluate the final evolved agent on BrowseComp-en and BrowseComp-zh without benchmark-specific zeroth-order trajectory discovery. As shown in Table~\ref{tab:additional_generalization}, our method improves over both the initial policy and RL on the two benchmarks. Together with the non-retrieval and non-verbatim analyses, these results show that the improvement is not limited to direct retrieval on the original evaluation benchmarks.

\begin{table}[t]
\centering
\footnotesize
\setlength{\tabcolsep}{3pt}
\renewcommand{\arraystretch}{0.94}
\caption{Results on non-retrieval and non-verbatim GAIA subsets and direct transfer to BrowseComp-en and BrowseComp-zh. The first two rows report accuracy, and the last two rows report Pass@8.}
\label{tab:additional_generalization}

\begin{tabular*}{\columnwidth}{@{\extracolsep{\fill}}lccc@{}}
\toprule
Evaluation & Initial & RL & Ours \\
\midrule

GAIA non-retrieval (56)
& 8.93 & 19.64 & \textbf{28.57} \\

GAIA non-verbatim (68)
& 10.29 & -- & \textbf{30.88} \\

BrowseComp-en
& 2.29 & 3.40 & \textbf{4.11} \\

BrowseComp-zh
& 17.64 & 21.45 & \textbf{23.53} \\

\bottomrule
\end{tabular*}
\end{table}

\subsubsection{Multiple random seeds}
We repeat the main controlled evaluation with five independent random seeds and report the mean, standard deviation, and 95\% confidence interval. As shown in Table~\ref{tab:multiseed_main_results}, the gains remain consistent across seeds. We also repeat the Pass@8 trajectory-discovery evaluation and obtain success rates of $63.5\pm0.4\%$ on the training set and $41.9\pm0.8\%$ on the validation set.

\begin{table}[t]
\centering
\footnotesize
\setlength{\tabcolsep}{3.5pt}
\renewcommand{\arraystretch}{0.92}
\caption{Main results over five independent random seeds. Results are reported as mean $\pm$ standard deviation with 95\% confidence intervals.}
\label{tab:multiseed_main_results}
\resizebox{\columnwidth}{!}{
\begin{tabular}{lcc}
\toprule
Method & GAIA (\%) & WebWalkerQA (\%) \\
\midrule
Initial agent
& $19.1\pm1.9$ $(16.8,21.5)$
& $19.7\pm1.9$ $(17.3,22.0)$ \\
Ours
& $\mathbf{28.6\pm0.6}$ $\mathbf{(27.7,29.4)}$
& $\mathbf{28.3\pm0.8}$ $\mathbf{(27.3,29.3)}$ \\
\bottomrule
\end{tabular}
}
\end{table}

\subsection{Efficiency}

We evaluate the efficiency of the parallel perturbation inference and adaptive lookup mechanism.

\paragraph{Parallel perturbation inference.}
The maximum trajectory context length is fixed to 20k tokens. We set $K=2$ and $K=4$ for comparison.
We measure the wall-clock time of unperturbed rollout, perturbed rollout sampling without the parallel mechanism, and the perturbed rollout sampling with our parallel mechanism.  Results are shown in Table~\ref{tab:parallel_efficiency}. 

Results show that parallel perturbation inference reduces the required computational time. Although perturbed sampling is still more expensive than a single unperturbed rollout, the extra cost is much smaller than running each perturbation independently. This makes multi-direction zeroth-order estimation feasible under limited hardware.



\begin{table}[t]
\centering
\footnotesize
\setlength{\tabcolsep}{3.5pt}
\renewcommand{\arraystretch}{0.92}
\caption{Efficiency of parallel perturbation inference. Sequential time is estimated by multiplying the single-rollout time by $K$.}
\label{tab:parallel_efficiency}
{
\begin{tabular}{ccccc}
\toprule
$K$ & Single Rollout & Seq. Estimate & Parallel Time & Reduction \\
\midrule
2 & 164s & 328s & 232s & 29.3\% \\
4 & 164s & 656s & 305s & 53.5\% \\
\bottomrule
\end{tabular}
}
\end{table}

\paragraph{Adaptive lookup mechanism.}
We then analyze the time saved by the adaptive lookup mechanism. Since different perturbations often lead to similar \texttt{search} queries or visit overlapping webpages, the lookup pool can reuse previous tool observations. 
We report the cache hit counts for \texttt{search} and \texttt{visit} under different perturbation numbers and optimization rounds.
We estimate the saved time using an average latency of 3 seconds for each \texttt{search} call and 10 seconds for each \texttt{visit} call. Results are shown in Table~\ref{tab:lookup_efficiency}. Results indicate that adaptive lookup reduces redundant external interactions while preserving exploration.



\begin{table}[t]
\centering
\footnotesize
\setlength{\tabcolsep}{3.5pt}
\renewcommand{\arraystretch}{0.92}
\caption{Cache hits and estimated saved time from the adaptive lookup mechanism.}
\label{tab:lookup_efficiency}

\begin{tabular*}{\columnwidth}{@{\extracolsep{\fill}}ccccc@{}}
\toprule
$K$ & Rounds & Search Hits & Visit Hits & Saved Time \\
\midrule
2 & 15 & 46 & 27 & 408s \\
2 & 30 & 78 & 45 & 684s \\
4 & 15 & 86 & 50 & 758s \\
4 & 30 & 150 & 84 & 1290s \\
\bottomrule
\end{tabular*}

\end{table}

\section{Conclusion}

We have presented a zeroth-order optimization method for self-evolving LLM agents that enables agents to learn beyond their capability boundaries by perturbing LLM parameters to fit difficult examples. 
The proposed method computes the losses under both perturbed and original parameters, uses the loss difference to estimate gradients, and updates the parameters accordingly. 
The updated LLM agent is then used to sample trajectories for supervised fine-tuning, forming a closed self-evolution loop. 
The introduced adaptive lookup mechanism and parallel perturbation inference mechanism reduce the time cost of zeroth-order optimization, while the proposed answer perplexity loss provides smooth and stable loss values for zeroth-order gradient estimation. 
Experiments on multiple deep research benchmarks demonstrate that our method obtains substantially more successful trajectories and consistently outperforms strong baselines. 
The comparison with existing methods demonstrates that our method can break through the inherent capability boundary of the agent itself.

\section*{Limitations}
Currently, we only demonstrate the effectiveness of our method on LLM agents, without extending it to the multimodal domain. 
In future work, we will further validate the superiority of our method on multimodal agents, following by~\citep{gao2025multi,li2026iterative,gao2024clova}.


\FloatBarrier
\bibliography{custom}

\appendix

\section{Responsible Research Considerations}
\label{app:responsible}

\subsection{Potential Risks}
\label{app:risks}

This work aims to improve the self-evolution capability of LLM agents in deep research settings. Although our experiments are conducted in controlled research settings on public benchmarks, stronger autonomous agents may introduce potential risks. First, agents may retrieve, synthesize, or propagate inaccurate information from external sources. Second, improved tool-use ability may lead to over-reliance on external search or webpage content without sufficient verification. Third, such agents could potentially be misused for large-scale automated information gathering. We mitigate these risks by focusing on benchmark-based evaluation, using controlled tool environments, and emphasizing that released artifacts are intended for research use.

\subsection{Scientific Artifacts}
\label{app:artifacts}

This work uses existing scientific artifacts, including public benchmarks, training data, open-weight model backbones, and baseline methods, for research purposes. We cite the creators of the artifacts used in our experiments, including GAIA, WebWalkerQA, WebShaper, ReAct, Qwen backbones, and related deep-research agent baselines. We use these artifacts in a manner consistent with their intended research use: benchmarks are used for evaluation, training data is used for research training and analysis, and baseline methods are used for comparison.

We respect the licenses and terms of use of the artifacts used in this work. For any artifacts created by this work, including sampled trajectories and trained agents, we will release them with appropriate documentation and under access and distribution conditions compatible with the licenses and terms of the original artifacts. In particular, derivatives of data accessed for research purposes will not be distributed outside research contexts when the original access conditions impose such restrictions.

\subsection{Data Privacy and Offensive Content}
\label{app:privacy}

This work does not intentionally collect personally identifying information or offensive content. The experiments are conducted using public research benchmarks and open research datasets. The generated trajectories are produced for research purposes in controlled agent environments.

Since deep-research agents may retrieve information from external web sources during tool use, generated trajectories may occasionally contain sensitive, personally identifying, or offensive content from retrieved webpages. Before releasing any generated trajectories or trained agents, we will conduct filtering and inspection to reduce such risks. This includes removing or anonymizing content that directly identifies private individuals when it is not necessary for the task, and excluding trajectories that contain offensive or harmful content unrelated to the benchmark question. Released artifacts will be accompanied by documentation describing their intended research use and any known limitations.

\subsection{Artifact Documentation and Data Statistics}
\label{app:artifact-documentation}

This work uses and creates scientific artifacts for research on deep-research LLM agents. The existing artifacts used in our experiments include public benchmarks, training data, open-weight model backbones, and baseline methods. The main evaluation benchmarks are GAIA and WebWalkerQA. GAIA evaluates general assistant capabilities such as reasoning, browsing, multimodal handling, and tool use. WebWalkerQA focuses on website traversal and multi-page information extraction. The training data is constructed from WebShaper QA pairs, which are used for research training and trajectory discovery in information-seeking tasks.

The artifacts are primarily task-oriented QA and agent-trajectory artifacts. They are not designed to represent demographic groups, and we do not use demographic annotations in our experiments. The work focuses on deep research, web traversal, information seeking, reasoning, and tool-use phenomena. The released artifacts created by this work may include sampled high-quality trajectories and trained agents, accompanied by documentation describing their intended research use, data source, filtering procedure, and known limitations.

For data statistics, we start from 500 QA pairs from the open-source WebShaper data and filter 304 high-quality QA pairs for training. In our trajectory analysis, the baseline model correctly answers 67 out of the 304 training examples under Pass@1 and fails on the remaining 237 examples. After training with the proposed zeroth-order optimization method, the optimized agent correctly answers 164 examples under Pass@1. We evaluate the trained agents on GAIA and WebWalkerQA following the official evaluation protocol.

\subsection{Computational Experiments and Implementation Details}
\label{app:computational}

We conduct computational experiments on deep-research LLM agents. The evaluation benchmarks include GAIA and WebWalkerQA, and the training data is constructed from 304 filtered high-quality QA pairs from WebShaper. We compare our method with ReAct under controlled settings and also report contextual comparisons with existing deep-research agents. We evaluate answer accuracy following the official evaluation protocol of each benchmark.

The main model backbones used in our controlled experiments are Qwen-3-4B and Qwen-3-8B. All controlled methods use the same backbone family and the same system configuration, including the search API, webpage parser, decoding setting, and tool budget.

For zeroth-order optimization, we optimize instance-specific LoRA modules by sampling random perturbation directions and estimating gradients from answer-loss differences. We evaluate perturbation numbers $K=2$ and $K=4$ in the efficiency analysis. The maximum trajectory context length is fixed to 20k tokens. The answer perplexity loss is used as the main loss for zeroth-order trajectory discovery. For supervised fine-tuning, sampled high-quality trajectories are used as training data for the self-evolved agent.

We use existing software packages and tools for model inference, LoRA fine-tuning, benchmark evaluation, search, and webpage parsing. Specifically, we use vLLM for LLM inference, Modified LLaMA-Factory for LoRA-based fine-tuning, and the official evaluation scripts or protocols of GAIA and WebWalkerQA for answer accuracy evaluation. We use Google Serper as the search tool and Jina.visit for webpage parsing. All methods in the controlled comparison use the same implementation settings for these tools.

The experiments were run on NVIDIA RTX3090 GPUs. The total computational budget was approximately 1152 GPU hours, including zeroth-order trajectory discovery, supervised fine-tuning, and benchmark evaluation. The main training and inference infrastructure consisted of 8 NVIDIA RTX3090 24GB GPUs. We will release additional implementation details with the code and artifacts.

We report benchmark accuracies as single-run results unless otherwise specified. For trajectory discovery, we report the number of successful examples under Pass@1. Among the 304 training examples, the baseline model correctly answers 67 examples and fails on 237 examples, while the optimized agent correctly answers 164 examples. For efficiency analysis, we report rollout wall-clock time, cache hit counts, and estimated saved time.

\subsection{Human Subjects, Consent, and Ethics Review}
\label{app:human-subjects}

This work does not involve human subjects, human participants, crowdworkers, or newly recruited human annotators. We do not conduct user studies, collect human responses, or ask human annotators to label new data. The experiments are conducted using existing public research benchmarks, open research datasets, automatic agent rollouts, and official evaluation protocols.

Because the work does not involve newly recruited participants or newly collected personal data requiring participant consent, data consent procedures are not applicable. Similarly, because there is no human-subject data collection protocol, ethics review board approval was not required for this work.

\subsection{Use of AI Assistants}
\label{app:ai-assistants}

The authors used AI assistants during the preparation of this paper. AI assistants were only used for language polishing. They were not used to generate scientific claims without author verification. All scientific ideas, experimental designs, analyses, results, and final claims were reviewed, edited, and verified by the authors.

\end{document}